\documentclass[a4paper,fleqn]{cas-dc}

\usepackage[numbers]{natbib}
\usepackage{microtype} 
\usepackage{amsmath}
\usepackage{flushend}
\usepackage{algorithm}
\usepackage[noend]{algpseudocode}
\usepackage{changepage} 
\makeatletter
\def\BState{\State\hskip-\ALG@thistlm}
\makeatother

\usepackage{array} 

\usepackage{subfiles}
\usepackage{comment}
\usepackage{todonotes}
\usepackage{scalerel}
\usepackage{booktabs}
\usepackage{multirow}
\usepackage{cite}
\usepackage{paralist}
\usepackage{float}

\def\tsc#1{\csdef{#1}{\textsc{\lowercase{#1}}\xspace}}
\tsc{WGM}
\tsc{QE}

\begin{document}
\let\WriteBookmarks\relax
\def\floatpagepagefraction{1}
\def\textpagefraction{.001}

\shorttitle{An Online Framework for Cognitive Load Assessment in Assembly Tasks}    

\shortauthors{Lagomarsino et al.}  

\title [mode = title]{An Online Framework for Cognitive Load Assessment\\in Assembly Tasks}  

\author[1,2]{Marta Lagomarsino}[type=editor,
    orcid=0000-0001-9121-1812]
\cormark[1]
\ead{marta.lagomarsino@iit.it}

\affiliation[1]{organization={Human-Robot Interfaces and Physical Interaction Laboratory, Istituto Italiano di Tecnologia},
            addressline={Via San Quirico 19d}, 
            city={Genoa},
            postcode={16163}, 
            country={Italy}
            }
\affiliation[2]{organization={Department of Electronics, Information and Bioengineering, Politecnico di Milano},
            addressline={Via Giuseppe Colombo, 40}, 
            city={Milan},
            postcode={20133}, 
            country={Italy}
            }        

\author[1]{Marta Lorenzini}
\author[2]{Elena De Momi}
\author[1]{Arash Ajoudani}

\cortext[1]{Corresponding author}

\begin{abstract}
The ongoing trend towards Industry 4.0 has revolutionised ordinary workplaces, profoundly changing the role played by humans in the production chain.
Research on ergonomics in industrial settings mainly focuses on reducing the operator's physical fatigue and discomfort to improve throughput and avoid safety hazards.
However, as the production complexity increases, the cognitive resources demand and mental workload could compromise the operator's performance and the efficiency of the shop floor workplace. 
State-of-the-art methods in cognitive science work offline and/or involve bulky equipment hardly deployable in industrial settings.
This paper presents a novel method for online assessment of cognitive load in manufacturing, primarily assembly, by detecting patterns in human motion directly from the input images of a stereo camera.
Head pose estimation and skeleton tracking are exploited to investigate the workers' attention and assess hyperactivity and unforeseen movements.
Pilot experiments suggest that our factor assessment tool provides significant insights into workers' mental workload, even confirmed by correlations with physiological and performance measurements.
According to data gathered in this study, a vision-based cognitive load assessment has the potential to be integrated into the development of mechatronic systems for improving cognitive ergonomics in manufacturing.
\end{abstract}


\maketitle

\section{Introduction}

\begin{figure*}[!t]
    \centering
    \includegraphics[width=\linewidth]{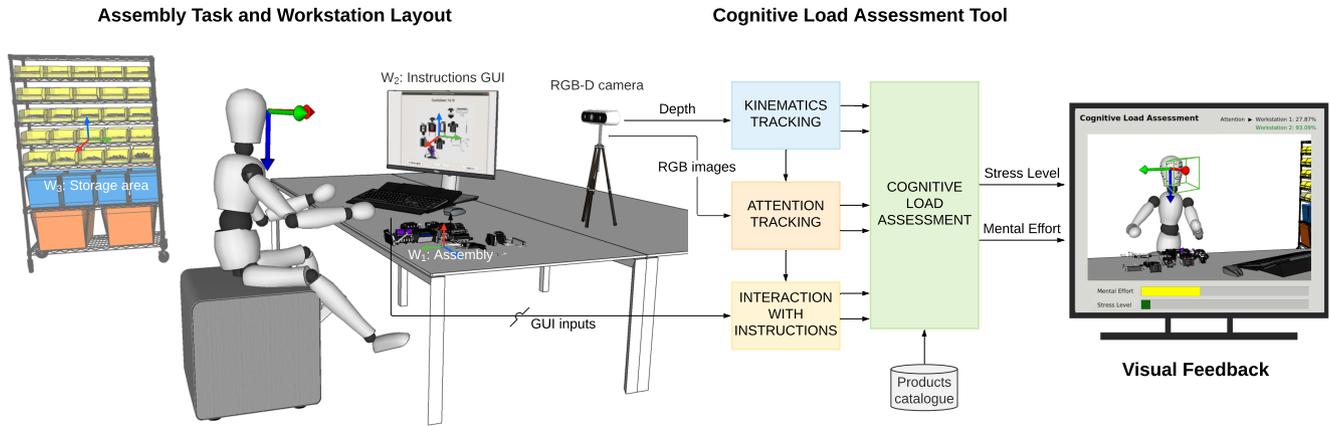}
    \caption{The system overview. Left: Conceptual illustration of workstation layout including: RGB-D camera, assembly, instructions graphical user interface (GUI), and storage area. Right: Block diagram of the proposed online framework to assess cognitive load and provide visual feedback to the user.} 
    \label{fig:CL_architecture}
\end{figure*}

Mental health problems at work affect hundreds of millions of people worldwide. About 17.6\% of the global working population suffer from common mental disorders (CMD) \citep{Steel2014}, such as anxiety, bipolarity and acute stress. The annual prevalence attains 38.2\% in the European Union, embracing attention-deficit hyperactivity disorder (ADHD), insomnia (7.0\%), and major depression (6.9\%) \citep{Wittchen2011}.
Many recent surveys \citep{Kubicek2019} 
and systematic reviews \citep{vanderMolene2020, Kayla2018} 
indicate the inadequate organisation and management of the work as a primary cause of such disorders and outline the relationship between excessive working pressures and demands and the incidence of depression, poor health functioning, anxiety, distress, fatigue, job dissatisfaction and burnout. 

Besides, work-related stress and psychological risks have direct financial implications for private companies and governments. 
In Europe, the cost related to mental illness symptoms is around 617 billion euros annually, including employers' expenses (absenteeism, presenteeism, turnover and loss in productivity) and social welfare costs \citep{Hassard2014}. 

On the other hand, the introduction of hybrid manufacturing systems, where workers and autonomous machines operate in close proximity, has contributed to changing the role of the human in the production chain, resulting in new occupational safety and health (OSH) challenges. 
The digitalisation of the actual workplace has led to work intensification, constant time pressure and adaptation to rapid and frequent changes in customer demand and requirements (i.e. goods to produce and services to offer).
Many of these changes provide development opportunities, nevertheless, they may perilously increase cognitive demand, when inadequately handled, and result in adverse health and safety hazards.
Consequently, the elevated mental workload may compromise the operator's performance and the efficiency of the workplace.

The study of human cognitive factors will supplement the well-established research on physical ergonomics \citep{Kim2017, Lorenzini2019}, to comprehensively understand how humans interact with the environment and facilitate a reduction of the workload. 
In addition, various studies have shown that psychological factors at work may have a significant influence on the development of musculoskeletal disorders (MSDs) \citep{Mehta2016}. 
For instance, mental workload, fatigue, and job stress can alter biomechanical control strategies for upper extremities (i.e. neck, shoulders, arms, and hands) and low back extension, as well as increase gait and sway variability \citep{Grobe2017}. 
As a final consequence, the phenomenon may induce muscle pain in the worker and even occupational injuries.

The global burden of work-related mental disorders is expected to increase year on year \citep{HSEstress2020} and can no longer be overlooked. 
Despite cognitive load theory has aroused much interest in the last decade \citep{Paas2003}, 
the study of cognitive load in manufacturing operations is a moderately new topic \citep{Carvalho2020, Gualtieri2021}. 
The field of \textit{Cognitive Manufacturing} \citep{IBM2017} (i.e. the usage of data across systems, equipment and processes to optimise the manufacturing performance) has only very recently aimed to attain information about human workload.

To the best of our knowledge, available tools can be used almost exclusively by experts or merely provide offline insights about the cognitive process (e.g. subjective questionnaires \citep{Valdehita2004}). 
A first attempt toward a more usable tool was made by Thorvald et al. \citep{Thorvald2019}, who developed an analytic method, denoted Cognitive Load Assessment for Manufacturing (CLAM), for assessing the cognitive burden that the worker is expected to employ within a particular assembly task and workstation layout. As a matter of fact, manual assembly is an essential activity in the manufacturing sector, which exposes workers to situations with varying cognitive demands \citep{Brolin2017}. When combining the latter with high time pressure, an increase in mental load frequently occurs \citep{Yabuki2017}. The tool is intended to be used directly by workers involved in the manufacturing domain. Nevertheless, such evaluation is still made offline, asking the end-users to fill a form and rate a set of factors associated with different aspects of their daily activity.

The scientific and industrial communities still need to be provided with a validated set of models and metrics for the cognitive workload. Particularly, gaps were identified in relation to the online assessment of the mental demand inflicted by manufacturing tasks. 

To respond to this challenge, the purpose of this paper is to develop a quantitative and online method to examine how industrial work affects people relative to their attention distribution, decision-making, mental overload, frustration, stress and errors.
We propose an online framework to monitor the cognitive workload of human operators by detecting patterns in their motion directly from the input images of a stereo camera. Head pose estimation and skeleton tracking are exploited to investigate the workers' attention and assess hyperactivity and unforeseen movements (see system overview in Figure \ref{fig:CL_architecture}). 
The developed tool computes a list of indicators associated with different aspects of an assembly task and workstation layout in manufacturing. Each factor impacts with a weight on two defined indexes: the \textit{mental effort} and psychological \textit{stress level}. 
According to the scores interval, we determine the level of cognitive load an individual is experiencing within the current setup.
The study employs assembly experiments to validate our online framework against state-of-the-art offline methods in the field of cognitive science (i.e. physiological signals, secondary task-performance measure and subjective questionnaires). 

The paper is structured as follows. 
In section \ref{cognitive_load} we characterise cognitive load and provide an overview of related works about the methods to measure it. 
Next, we present our framework for the online assessment of mental effort and stress level. 
Pilot experiments are then proposed in section \ref{experiments} and the result are discussed and validated through statistical analysis.
The final sections discuss the contributions and limitations of the framework.

\section{Related works}
\label{cognitive_load}

The evidence that undue cognitive demand at work can prejudice the mental health of workers and their manufacturing performance has increased the interest in cognitive load theory (CLT). 
CLT investigates the interaction of cognitive structures, information and its implications \citep{Sweller1998}. 
In particular, the term cognitive load refers to the amount of processing that performing a particular task imposes on the learner's cognitive system \citep{Paas2003}.
Xie and Salvendy \citep{Xie2000} present a detailed conceptual framework of human information processing and distinguish between \textit{instantaneous} and \textit{overall load}. 
\textit{Instantaneous load} is defined as the dynamics of cognitive load, which constantly fluctuates over time as a response to stimuli that the present activity and environmental conditions are imposing on the subject. 
\textit{Overall load} results by the whole working procedure and represents the experienced and garnered instantaneous load in the human's brain.

A large and growing body of literature has investigated techniques to model human mental workload \citep{Parasuraman2008} and quantify the cost of performing tasks \citep{Xie2000, Haji2015}. 
Paas and Van Merriënboer \citep{Paas2003} describe mental load, mental effort, performance, and level of stress as the measurable dimensions of cognitive load.
Generally, cognitive load measurements belong to three main categories: physiological measures, subjective rating scales and performance-based measures.

Physiological measurement of workload relies on evidence that increased mental demands lead to an increased physical response from the body \citep{Sweller1998}. 
Various researchers have investigated the relationship between mental effort and heart rate variability (HRV) metrics in three frequency bands of interest: very low frequency (VLF, 0–0.04 Hz), low frequency (LF, 0.04–0.15 Hz), and high frequency (HF, 0.15–0.4 Hz) \citep{Delliaux2019, Durantin2014}. 
According to recent studies, intense cognitive demand leads to a decrease in HF power and a growth in the LF, respectively related to a parasympathetic withdrawal and a predominant increase in sympathetic activity \citep{Delliaux2019, Mizuno2011}. 
Besides, the galvanic skin response (GSR, also known as electrodermal activity, EDA) has been widely studied to quantify cognitive states \citep{Setz2010}. GSR or EDA is the measure of the continuous changes in the skin's electrical conductance caused by the variation of the sweating activity of the human body. 
The signal is typically described as a combination of two components, the tonic and phasic response.
Researchers use high-resolution EDA for indexing variations in sympathetic arousal associated with emotion, cognition, and attention \citep{Marucci2021, Rajavenkatanarayanan2020} and today represents one of the preferred metrics for stress \citep{Kyriakou2019}.
More recent studies also include measures of respiratory activity \citep{Grassmann2016}, eye activity \citep{Coyne2016, DiStasi2016}, cortisol level \citep{Carrasco2003}, speech measures \citep{Yin2008}, 
and brain activity \citep{Rosanne2021}.

Psychophysiological measurements provide objective and quantitative information, as well as the possibility to visualise a continuous trend and identify detailed patterns of load. However, these signals are highly sensitive to human movements, and the sensory acquisition system may be bothersome for the users and condition normal activities, severely limiting the adoption in real-world scenarios. 

Thus far, the measurement of cognitive load in laboratory settings mainly relies on subjective rating scales \citep{Valdehita2004}. 
The most commonly used questionnaire is called NASA-Task Load Index (NASA-TLX) \citep{Hart1988}.
Self-ratings nevertheless have many limitations \citep{Naismith2015}. Firstly, they are based on the assumption that people are able to introspect on the cognitive processes and report the amount of experienced cognitive effort. Secondly, they are often affected by many biases, such as acquiescence and social desirability. 
Lastly, the data are delivered after the completion of the activity and can be exploited only following extensive analysis by experts in the area of cognitive ergonomics and cognitive science. 

The third alternative to measuring cognitive load is through task- and performance-based techniques. Various metrics are presented in the literature (e.g. reaction time, accuracy and error rate) to assess the performance of both the primary and secondary tasks \citep{Haji2015}. The secondary task is performed concurrently and is supposed to reflect the level of the cognitive load imposed by the primary task \citep{Paas2003}. 
Despite the high sensitivity and reliability, this technique can be rarely applied, even in laboratory settings.

All the studies reviewed here support the hypothesis that existing approaches for cognitive load assessment have their strength and weakness and can be sensitive to distinctive aspects of workload.
When measuring workload empirically, the rule of thumb is to select a variety of measurements that seem appropriate to the application and are likely to provide insights into cognitive processes \citep{Miller2001}.
Unfortunately, most of these techniques are potentially difficult to be applied in industrial scenarios. Indeed, they require rather expensive and impractical equipment that may be uncomfortable for the users.

Despite the increasing enthusiasm to understand the multidimensional construct of the mental workload, the cognitive manufacturing field is still looking for practical solutions \citep{Carvalho2020}. Our work responds to the growing need to gather online data giving insights about the mental processing system and enables the identification of excessive cognitive load of assembly workers.

\section{The Cognitive Load Assessment Framework}

\begin{figure*}[!b]
    \centering
    \includegraphics[width=\linewidth]{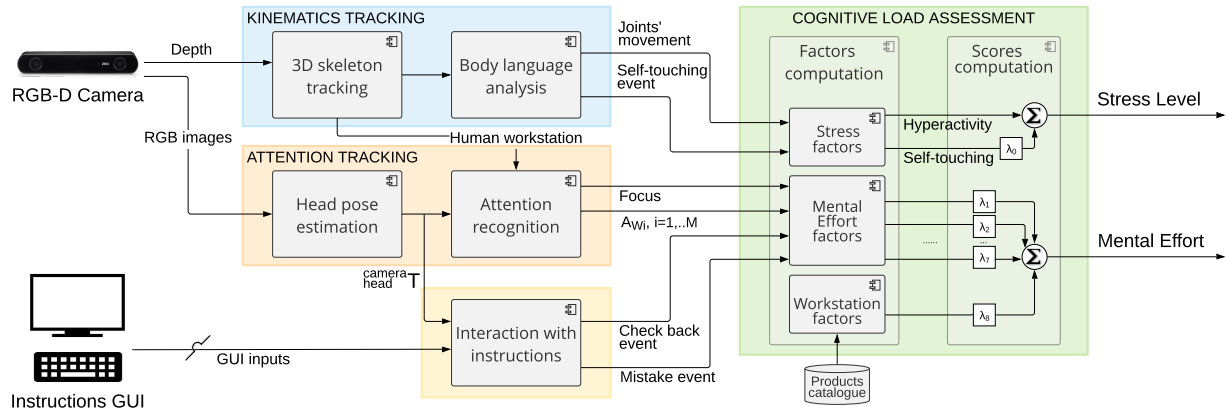}
    \caption{Overall structure of the online cognitive load assessment framework. 
    The proposed approach detects patterns in human's motion (blue block), investigates workers' attention (orange block) and their interaction with assembly instructions on a monitor (yellow block). Combining all these factors, final scores of \textit{mental effort} and psychological \textit{stress level} are computed (green block).}
    \label{fig:CL_schema}
\end{figure*}

The overall structure of the proposed cognitive assessment framework is represented in Figure \ref{fig:CL_schema}.
Our method investigates
\begin{inparaenum}[(i)]
    \item the concentration level of a worker by considering gaze direction and head pose, 
    \item the stress level, by analysing activity-related body language (i.e. self-touching occurrences and high activity periods) and 
    \item the information and part identification cost, namely the cognitive effort required to utilise the assembly instructions and handle the right tools and components to complete the task.
\end{inparaenum}
Additionally, we include \textit{a priori} defined parameters reflecting features of the specific assembly task and workstation layout (e.g. the number of assembly parts and noise level).
Combining all these factors, we compute the final scores of \textit{mental effort} and \textit{stress level}. This enables us to identify excessive cognitive load in the assembly workers. Besides, the framework includes a visual feedback interface, through which intuitive warning messages can be provided to the assemblers. 

To easily deploy the proposed framework in both laboratory and industrial settings, the choice of the external sensory systems was driven by the implementation costs and users' comfort (e.g. by avoiding wearability constraints). 
Hence, we selected a family of affordable active 3D imaging systems, namely RGB-D cameras, to detect human operators and quantify their workload.
The depth information and RGB images of the camera are processed by the `human upper-body kinematics tracking' module and the `attention tracking' module to compute a set of cognitive load factors introduced in Section \ref{CLfactors}. These two modules operate in synergy with the `interaction with instructions' module and converge into the `cognitive load assessment' module to compute the final scores of \textit{mental effort} and \textit{stress level} (see Figure \ref{fig:CL_schema}). 

Before describing the modules in detail, we provide the definition of the workstation layout. An operating environment can be defined by the involved workstations and their relative configuration.
We consider at least three types of workstations\footnote{Throughout this paper, the term `operating environment' (also known as `working area' or `workplace') refers to a place available to manufacturing personnel to carry out work. The `workstation' is instead a specific location, e.g. an assembly table, where employees perform specific tasks.} in industrial assembly tasks: the \textit{assembly workstation} $W_1$, which is the area occupied by the assembly components, the \textit{instructions workstation} $W_2$, where assembly information are displayed through e.g. a monitor, 
and the \textit{storage area} $W_3$, where the assembly components (e.g. screws, nuts and tools) are stored.
Based on the number of workstations, the system accordingly associates reference frames (see Figure \ref{fig:CL_architecture}) in the position specified during a configuration phase. The locations of the reference frames with respect to the operator’s head are used to determine the level of attention toward every workstation 
(see Section \ref{sec:3C}). 

\subsection{Human Upper-Body Kinematics Tracking \\Module}
\label{sec:3B}

The central role of this module is to detect the presence of a human operator entering the working area and to provide information to the system about the variations of his/her kinematic body configuration over time. 
We exploit a visual skeleton tracking algorithm, developed by StereoLabs\footnote{\texttt{https://github.com/stereolabs}} to track the human skeleton from the input images of a stereo camera. 
The module is, however, scalable to any other visual tracking method, e.g. OpenPose \citep{Cao2009}, or even IMU-based motion capture systems, such as Xsens suit\footnote{\texttt{https://www.xsens.com/motion-capture}}. 
The algorithm extracts the 3D position of twenty-five human keypoints (e.g. neck, shoulders, elbows, wrists, hips, knees, ankles) in real-time. Among them, we select the ones belonging to the upper body, and we analyse their displacements to compute factors defining the operator \textit{stress level} (see Section \ref{stress}).

Spatio-temporal information of human movements is also used to distinguish between possible tasks performed by the operator. 
To do this, the distance on the horizontal plane between the "neck" skeleton keypoint and the workstations is continuously computed: the worker is assumed to perform the task associated with the workstation he/she is closest to.
For instance, we assume that the assembler is searching for a tool if he/she accesses the storage area.
On the other hand, the \textit{mental effort} factors (defined in Section \ref{sec:mental_effort_factors}) are computed only if the subject is within a predefined range with respect to the assembly or instruction workstation (i.e. $\textit{W}_1$ or $\textit{W}_2$). 

\subsection{Human Attention Tracking Module}
\label{sec:3C}

Nowadays, several sensory systems can provide accurate measurements of human engagement and attention, such as eye-tracking screen-based devices (e.g. Gazepoint GP3 \citep{Coyne2016}) 
and glasses (e.g. Tobii Glasses 2 \citep{DiStasi2016}), 
or electroencephalography headsets (e.g. Neurolectrics Enobio \citep{Rosanne2021}). However, these systems bring about significant disadvantages such as discomfort (in wearable systems) and limited operational range (in screen-based eye-tracking devices). For these reasons, we developed a vision-based module, which is briefly outlined in Algorithm \ref{code:attention}.  We exploit a head tracker\footnote{\texttt{https://github.com/yinguobing/head-pose-estimation}}, which adopts OpenCV to detect the human face and a TensorFlow pre-trained deep learning model to identify facial landmarks.  
To estimate the head pose, a Perspective-n-Point (PnP) problem between the OpenFace\footnote{\texttt{https://cmusatyalab.github.io/openface}} 3D model and the detector's output (i.e. sixty-eight keypoints in pixel coordinates) is solved using the OpenCV function \texttt{solvePnP}. The PnP problem is stated as an iterative method based on a Levenberg-Marquardt optimisation \citep{Levenberg1944} 
and the solution is the pose that minimises the reprojection error, namely the sum of squared distances between the observed projections on the image plane and the projected 3D points in the model.
A Kalman Filter is used to stabilise the pose computed frame by frame.

\begin{algorithm}[!t]
\small
\caption{Human Attention Tracking}\label{code:attention}
\begin{algorithmic}[1]
\Procedure{}{}
\State $^{\text{camera}}\textit{workstation} \gets \text{position of }\textit{W}_{1,2..M} \text{ w.r.t. camera}$
\State $\textit{face\_model} \gets \text{3D model of the face}$

\BState \emph{top}:
\State $\textit{img} \gets \text{new acquired image at time }t$
\State $\textit{face\_marks} \gets \text{detect\_face\&landmarks} (\,\textit{img}\,)$
\State $\textit{head\_pose} \gets \texttt{cv.solvePnP} (\,\textit{face\_model}, \textit{face\_marks}\,)$
\State $\textit{steady\_head\_pose}  \gets \text{kalman\_filter} (\,\textit{head\_pose}\,)$
\State $^{\text{camera}}_{\text{head}}T \gets \text{pose2TF} (\,\textit{steady\_head\_pose}\,)$
    
\BState \emph{loop}:
\For{\textbf{each} workstation $\textit{W}_i$}
    \State $^{\text{head}}\textit{workstation(i)} \gets  ^{\text{camera}}_{\text{head}}T^{\text{-1}} * \, ^{\text{camera}}\textit{workstation(i)} $
    \State $\theta_i, \ \varphi_i\ \gets \ \text{cartesian2spherical} (\,^{\text{head}}\textit{workstation(i)}\,)$
    \State $A_{\textit{W}i} \gets f(\theta_i) * f(\varphi_i)$
    \State $\textit{attention\_W(i)} \gets 0$
    \If {human $\in$ $\textit{W}_1 \lor \textit{W}_2$}
        \If {$A_{\textit{W}i} > \textit{threshold}$}
            \State $\textit{attention\_W(i)} \gets A_{\textit{W}i}$
        \EndIf
    \EndIf
\EndFor
\State $\textit{focus} = \text{arg\,max}_{\not{0}}(\,\textit{attention\_W(i)}\,)$

\State \textbf{goto} \emph{top}.
\EndProcedure
\end{algorithmic}
\footnotesize{Note that in $^ax$ the apex $a$ represents the reference frame in which the variable $x$ is expressed.}
\end{algorithm}

The output of the procedure is the location and orientation of the head with respect to the camera frame.
According to the estimated odometry, a frame is associated with the head and the transformation $^{\text{camera}}_{\text{head}}T$ expresses the head pose variation over time. 
Subsequently, we look up the transformation between the head frame and each workstation defined in the configuration phase and the Cartesian vector expressing their relative position is mapped in spherical coordinates (i.e. azimuth angle $\theta$, elevation angle $\varphi$ and radial distance).

A fuzzy logic membership function was modelled to estimate the level of attention toward each workstation. In particular, the azimuth and elevation values are separately transformed using a Raised-Cosine Filter \citep{Glover2004}, where a sigmoid normalises the values to a scale from zero to one. To obtain the desired behaviour in different ranges, we define the function as follows:

\begin{equation}
f(\alpha_i) =
\begin{cases}
    1, & \mbox{if } \left\lvert\alpha_i\right\rvert \leq\alpha_{\text{min}} \\
    \frac{1}{2} \bigl[ 1-\cos{\bigl(\frac{\left\lvert\alpha_i\right\rvert - \alpha_{\text{min}}}{\alpha_{\text{max}} - \alpha_{\text{min}}}\pi\bigr)} \bigr], & \mbox{if } \alpha_{\text{min}} < \left\lvert\alpha_i\right\rvert \leq\alpha_{\text{max}} \\
    0, & \mbox{if } \left\lvert\alpha_i\right\rvert > \alpha_{\text{max}}
\end{cases}
\end{equation}
where $\alpha_i$ is one of the two measured angles (i.e. azimuth $\theta_i(t)$ or elevation $\varphi_i(t)$) at each time instant $t$, allowing the continuous localisation of the $i$-th workstation with respect to the subject's gaze direction. The fuzzy function includes control points ($\alpha_{min},\alpha_{max}>0$) defined a priori, which determine the independent upper and lower limits of the area where the function has a smooth behaviour. 
Thus, the indicator $f(\alpha_i)$ decreases exponentially along with the growth of absolute angle value above the minimum threshold ($\alpha_{min}$), before levelling off at a maximum threshold ($\alpha_{max}$).

The assessment of the attention level $A_{\textit{W}i}$ toward each $i$-th workstation is therefore computed as the product between the normalised azimuth $\theta_i(t)$ and elevation $\varphi_i(t)$ indicators:
\begin{equation}
    A_{\textit{W}i}\bigl(\theta_i(t), \varphi_i(t)\bigl) =  f\bigl(\theta_i(t)\bigl) * f\bigl(\varphi_i(t)\bigl)
\end{equation}
Given the estimated attention to all workstations, we can assess if the worker is currently distracted or concentrated on a particular workstation. 
This is determined by simply checking if at least one of the attention parameters is above a predefined threshold. If it would be the case, we find out the workstation that the worker is looking at as the one in which the associated parameter $A_{\textit{W}i}$ is maximum. 

\subsection{Interaction with Instructions Module}
In this work, we assume that assembly instructions are shown on a monitor through a Graphic User Interface (GUI), permitting the operator to browse them (see Figure \ref{fig:CL_architecture}). 
Inputs from the keyboard permit to watch the next instruction, check the same instruction again (i.e. instruction check back) or go back in instructions.
As a consequence, the `interaction with instructions' module is in charge of monitoring the task advancement. According to registered keyboard commands, it provides the system with the number of steps of the assembly sequence that the user has already followed, the instruction check backs and the occurrence of an error in the assembly sequence that obliges the user to go back to more than one instruction. 

\subsection{Cognitive Load Assessment Module}
The last module exploits workload indicators in manufacturing as identified by several experienced researchers and industrial experts \citep{Thorvald2019}.
Particularly, we define a list of cognitive load factors and compute them starting from the output of the modules described above.
Note that the unit of analysis is on the workplace level, including both the operator and the workstations layout. 
Each factor is then multiplied by its assigned weight $\lambda$ (see Section \ref{experiments}.D), and a definite sum of the weighted metrics determines the final scores of \textit{mental effort} and \textit{stress level}.
A detailed description of the proposed cognitive load factors and scores can be found in the next section.

\section{Definition of cognitive load factors and \\final scores}
\label{CLfactors}

We define and develop a set of cognitive load factors that are computed for each system pipeline loop and contribute specifically to one of the aforementioned indexes (i.e. \textit{mental effort} and \textit{stress level}). 
Some of the factors include both an \textit{instantaneous} and \textit{overall} parameter, based on the cognitive load definitions provided at the beginning of Section \ref{cognitive_load}, and their specific usage will be explained afterwards.
In addition, we present `workstation factors', which may affect the total cognitive load in assembly tasks.
Note that the proposed indexes analyse the assembler behaviour within a predefined workstation layout.
Moreover, each factor is not expected to directly reflect human cognitive processing. Our position is that a combination of those factors could provide insights into the human cognitive system.

\subsection{Mental effort factors}
\label{sec:mental_effort_factors}

\subsubsection{Concentration Loss: }
This factor analyses the attention that an individual gives to a task. It is based upon contemporary psychology claim that cognitive load usurps executive resources, which otherwise could be used for attentional control, thus increases distraction \citep{Lavie2004}. 

Accordingly, we assess here the amount of time not dedicated to the assembly, instructions or any other defined workstation, and hence quantify how long an individual is not concentrated on his/her assembly task. The \textit{Concentration Loss} factor is thus defined as 
\begin{equation}
\small
    \textit{Concentration Loss}(t) = 1 - \sum_{w = 1}^{M}\frac{\text{[attention time]}_w}{\text{time elapsed}},
\end{equation}

\noindent where $M$ is the number of workstations defined in the configuration phase. The `$\text{[attention time]}_w$' is the interval in which the subject is focused on the $w$-th workstation $W_w$, namely $A_{Ww}$ is above the predefined threshold and $A_{Ww}>A_{Wi},\ \forall \ i\neq w$. 
For instance, `$\text{[attention time]}_2$' represents the time spent looking at the instructions on the monitor. 
Finally, the `time elapsed' refers to the time passed since the task starts, expressed in seconds\footnote{Note that `time elapsed' is defined in the same way for all the factors.}.

\subsubsection{Learning Delay: } 
This metric investigates the ability to rapidly learning a novel rule from instructions and assesses the operator's automaticity in completing the assembly.
We took inspiration from Rapid Instructed Task Learning \citep{Liefooghe2012, Cole2012} theory, which analyses the efficient action execution immediately following instructions and without prior practice. The studies highlight that instructions can even produce automatic effects in relatively simple tasks.

The assumption here is that the more time the subject spends focusing on the assembly components, the slower is the learning. Hence, we can infer that the less trivial is the task,  the higher is the cognitive load.
The \textit{Learning Delay} factor is thus defined as 
\begin{equation}
\small
    \textit{Learning Delay}(t) = \frac{\text{assembly time}}{\text{time elapsed}},
\end{equation}

\noindent where `assembly time' or `$\text{[attention time]}_1$' is the interval in which the subject focuses on the assembly workstation $W_1$. 

\subsubsection{Concentration Demand: } 
The estimated incidence of attention failures is usually associated with cognition overload \citep{Head2014}.
This factor is defined as the number of times the subject gets distracted, losing their attention toward all workstations involved in the task. 

In particular, the \textit{instantaneous} parameter evaluates the transitions to not attention per instruction, excluding the ones to shift the focus to another workstation, thus is defined as 
\begin{align}
\small \hspace{-0.9cm}
    \textit{Concentration Demand}(i_n) & \small = \sum_{w = 1}^{M}\bigl(\text{[losses of attention]$_w$} \nonumber \\
    & \small - \text{[switches to another workstation]$_w$}\bigl), 
\end{align}

\noindent where $i_n$ represents the $n$-th instruction and $M$ is the number of defined workstations.

The \textit{overall} parameter keeps the memory of load that the operator experiences during the task. Whenever the event $d$ (i.e. loss of attention from any workstation) is detected, we record the instant in which it occurs. Then, the ratio of the sum of the time instances and the time elapsed is considered:
\begin{equation}
\small \hspace{-0.9cm}
    \textit{Concentration Demand}(t) =
    \sum_{d=1}^{D} \frac{\text{[instant of attention loss}]_{d}}{\text{time elapsed}},
\end{equation}
\noindent where $D$ is the number of total occurrences of attention loss while working on the task.   
Note that each occurrence equally impacts the indicator, and as time passes, the contribution of a past event decreases.

\subsubsection{Instructions Cost: }
This metric examines the general quality of the instructions used to gather information about the work. The analysis relies on human-computer interaction guidelines and studies on the required cognitive effort to utilise them \citep{Chandler1991}. 
We counted the attention switches between the assembly workstation and the monitor, excluding the required checks for a new instruction.
The \textit{instantaneous} parameter defines the cost of information per instruction as
\begin{equation}
\small
    \textit{Instructions Cost}(i_n) = \text{instruction checks} - 1,
\end{equation}

\noindent where $i_n$ is the $n$-th instruction.
On the other hand, the \textit{overall} parameter considers the $C$ instants in which the event $c$ (i.e. a not required switch) occurred:
\begin{align*}
\small
    \text{not required switches} & \small = \text{instruction checks} \nonumber \\ 
    & \small - \text{instructions showed},
\end{align*}
\vspace{-2.6em}\\
\begin{equation}
\small
   \textit{Instructions Cost}(t) =
    \sum_{c=1}^{C} \frac{\text{[instant of not required switch]}_c}{\text{time elapsed}}
\end{equation}

\subsubsection{Task Difficulty: }
This factor estimates the required cognitive effort to perform a task. 
To do that, the framework automatically records the instructional check backs $b$ on the GUI. 
Since task demand can vary as a function of the cognitive load \citep{Klingner2011}, the \textit{instantaneous} parameter is also here complemented with an \textit{overall} parameter. The latters are thus defined as
\begin{align*}
\small
    \textit{Task Difficulty}(i_N) = \text{instruction check backs},
\end{align*}
\vspace{-2.6em}\\
\begin{equation}
\small
    \textit{Task Difficulty}(t) =
    \sum_{b=1}^{B} \frac{\text{[instant of instruction check back]}_b}{\text{time elapsed}}
\end{equation}
\noindent where $B$ is the total amount of instruction check backs performed during the task.

\subsubsection{Frustration by Failure: }
This is a simple metric describing the mechanism triggered after making a mistake $e$. The \textit{instantaneous} and \textit{overall} parameters are computed as for previous factors:
\begin{align*}
\small
    \textit{Frustration by Failure}(i_n) = \text{number of mistakes made}, 
\end{align*}
\vspace{-2.8em}\\
\begin{equation}
\small
    \textit{Frustration by Failure}(t) =
    \sum_{e=1}^{E} \frac{\text{[instant of mistake occurrence]}_e}{\text{time elapsed}}
\end{equation}
\noindent with $E$ the total amount of mistakes made during the task.
Here, an error in the assembly sequence is detected whenever the user goes back to more than one instruction. \newline

It should be reminded that, thanks to skeleton tracking, we detect in which workstation the operator is. Hence, please note that the factors described above are computed only if the human is in proximity to workstation $\textit{W}_1$ and remain constant if the operator moves away. 

\subsubsection{Tool Identification: }
This factor assesses the mental processing to identify the tool needed for the assembly.
Whenever the storage area is accessed (i.e. the human is in proximity to workstation $\textit{W}_3$), the \textit{Tool Identification} factor is computed as the time spent to seek the right tool in tenths of a second.

\subsection{Stress level factors}
\label{stress} 
The analysis of body language is gaining an increasing interest in the emerging field of automatic detection of stress \citep{Carneiro2012}.
Accordingly, we defined activity-related features solely based on visual information and the derived skeleton tracking. 

\subsubsection{Self-touching: }
It has been proven that self-touching is a behavioural indicator of stress and anxiety \citep{Harrigan1985}. 
We compute the distance between each hand and the head key points of the detected skeleton. If the value is below a predefined threshold, a self-touching occurrence is registered and impacts the final score for a minute:
\begin{equation}
\small
    \textit{Self-touching}(t) =
    \sum_{s=1}^{S} \frac{\text{[instant of self-touching]}_s+60-t}{\text{60}}
\end{equation}

\noindent where $S$ is the number of self-touching occurrences $\epsilon \ [t-60,t]$ and $t$ is the time in seconds.

\subsubsection{Hyperactivity: } 
An analysis of human motion is performed to detect stress-related high activity periods. Our method is solely based on visual spatiotemporal information of human kinematics extracted from video sequences representing the monitored subject.

During an initial calibration phase, we capture the upper body joints movement within two subsequent frames and store the mean $\mu_1,\mu_2... \mu_N$ and standard deviation  $\sigma_1,\sigma_2... \sigma_N$ (where $N$ is the number of selected upper body joints) of the sum of the displacements in a time window $\tau$. This baseline recording permits us to compare the online data with a session under resting conditions. 
Then, during task execution, we periodically compute the deviation of every joint from its mean motion. If a $j$-th joint's deviation $\Delta_j$ is greater than the stored standard deviation $\sigma_j$, a parameter associated with $j$-th joint, called `activity', is evaluated as the ratio between $\Delta_j$ and $\sigma_j$.
A unique descriptor of activity level is computed as the mean of all the upper body joints' activity.

\subsection{Workstation factors}

The worker can navigate the list of products and combine different objects in sequences to handle more complex assemblies. In the catalogue (.csv file), the number of components and required tools for each object are specified. 
The sequence of objects to assemble is loaded and the following parameters are evaluated for the selected task.

\subsubsection{Number of assembly components: }
A parameter, normalised between 0 and 1, rising linearly with the number of parts intended to be assembled into a complex product.

\subsubsection{Number of tools used: }
A normalised parameter describing the number of tool used to complete the assembly.

\subsubsection{Physical effort: }
The required physical effort to perform a task. 
The estimated difficulty factor takes values between 0 (simple - not previous experience is required) and 1 (difficult - significant training and experience are required).

\subsubsection{Variant flora: } 
An estimation of the level of variation on a workstation (from no variation, i.e. one-piece production, to full variation, i.e. flexible and customised production).\newline

In addition, several environmental factors such as lighting conditions, temperature and level of noise may influence the operators' conditions. While the first two can be considered rather constant in an industrial workplace, the level of noise may greatly vary depending on the working scenario. There is increasing evidence that chronic noise stress impairs cognition and induces oxidative stress in the brain \citep{Subramaniam2019}. 
With this in mind, the \textit{Level of noise} factor has been defined.
\subsubsection{Level of noise: }
The sound pressure level in manufacturing environment. 
A sound sensor could measure the surrounding ambient sound in the audible frequency spectrum for the human ear. 
Given the mean level of noise $\mu_{noise}$ in A-weighted decibels (dBA), the parameter is defined as follows: 
\begin{equation}
\small
    \textit{Level of noise} = 
    \begin{cases} 
    0,  & \mbox{if } \mu_{\text{noise}}\leq20 \mbox{ dBA} \\ 
    \mbox{parabolic function},  & \mbox{if } 20<\mu_{\text{noise}}\leq70 \mbox{ dBA} \\
    1,  & \mbox{if } \mu_{\text{noise}}>70 \mbox{ dBA}
    \end{cases}
\end{equation}

\noindent where the thresholds (20 and 70 dBA) are defined in compliance with recommended standard occupation noise exposure \citep{Subramaniam2019}. 

\subsection{Cognitive load scores estimation}
\label{online_framework}

The cognitive load factors described in the previous sections are computed online, in the background of workers normal activities. 
This is to identify excessive cognitive load on the fly and deliver warning messages to the assembly worker. 
With this aim, we multiply each factor by a weight (see Section \ref{experiments}.D): the sum of the weighted \textit{mental effort} and \textit{stress level}
factors results respectively in the two homonyms `higher-level' scores. 

The \textit{mental effort} is computed at two different levels. Its dynamics is estimated online exploiting the \textit{instantaneous} parameters and provided as feedback through a dedicated screen.
A detailed description of the visual feedback interface is presented in Section \ref{GUI}.
For the post hoc analysis, we instead select the \textit{overall} factors since, at this stage, we do not aim to evaluate the cognitive load triggered by a stimulus but the overall mental effort induced by the whole task and cross-compare its trend among diverse testing conditions. 

On the other hand, the \textit{stress level} score is defined by the hyperactivity, plus each occurrence of self-touching impacts with a predefined value on the final score, and, as time passes, its contribution decreases.

\section{Experimental analysis}
\label{experiments}
In this section, the experimental campaign to validate our framework is described in detail. We adopted both quantitative and qualitative measures to assess the performance and potentials of the proposed approach. 

\begin{figure}[!t]
    \centering
    \includegraphics[width=\linewidth]{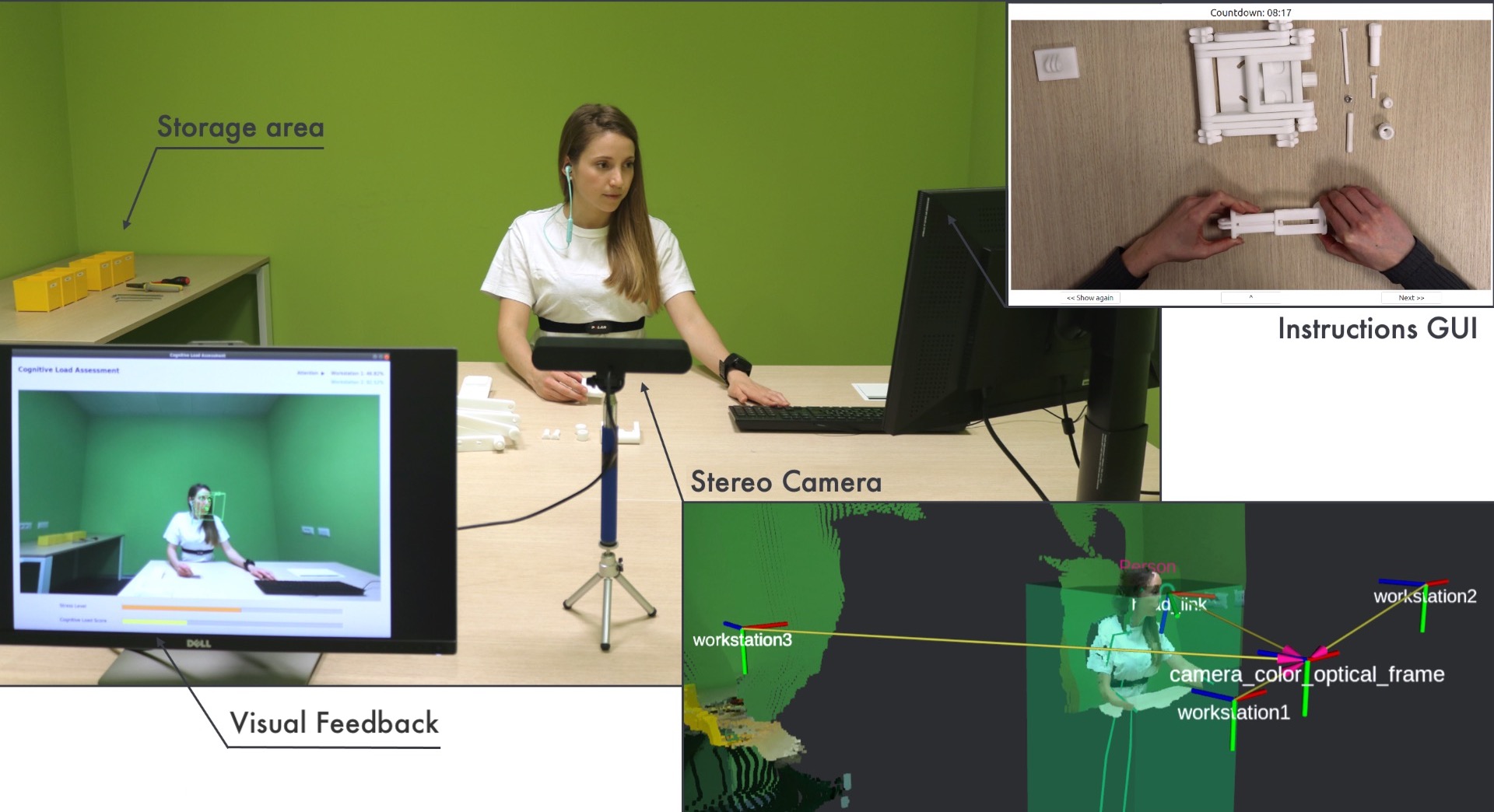}
    \caption{Overview of the experimental setup highlighting: zed2 stereo camera, instructions GUI (monitor and keyboard), storage area and screen providing visual feedback. The 3D assembly is placed on the table in front of the subject.}
    \label{fig:experimentalSetup}
\end{figure}

\subsection{Experimental setup} 
For the experiments, we reproduced a possible operating environment in our laboratory (see Figure \ref{fig:experimentalSetup}). 
The participants were asked to sit at a desk, and a 3D printed assembly kit\footnote{\texttt{https://tinyurl.com/3Dprintedassembly}} was placed on the table (defining workstation $\textit{W}_1$).
The instructions to assemble the object were shown on a monitor (workstation $\textit{W}_2$) and consisted of short videos of about $20$ s each. The user could browse them through a GUI. Inputs from the keyboard permitted to watch the next instruction, reproduce the same instruction video (i.e. instruction check back) or go back in instructions.
Finally, small boxes with screws, bolts, nuts and required tools were placed in the area right behind the participant defining workstation $\textit{W}_3$.

A stereo camera (zed2, Stereolabs, San Francisco, CA, USA) monitored the participant from the front for the entire duration of the experiment. Note that the framework does not require the recording of a video (i.e. the computations were performed online), however, it was acquired as a backup to measure the detection accuracy of subjects' motion patterns.

\begin{figure}[!b]
    \centering
    \includegraphics[width=\linewidth]{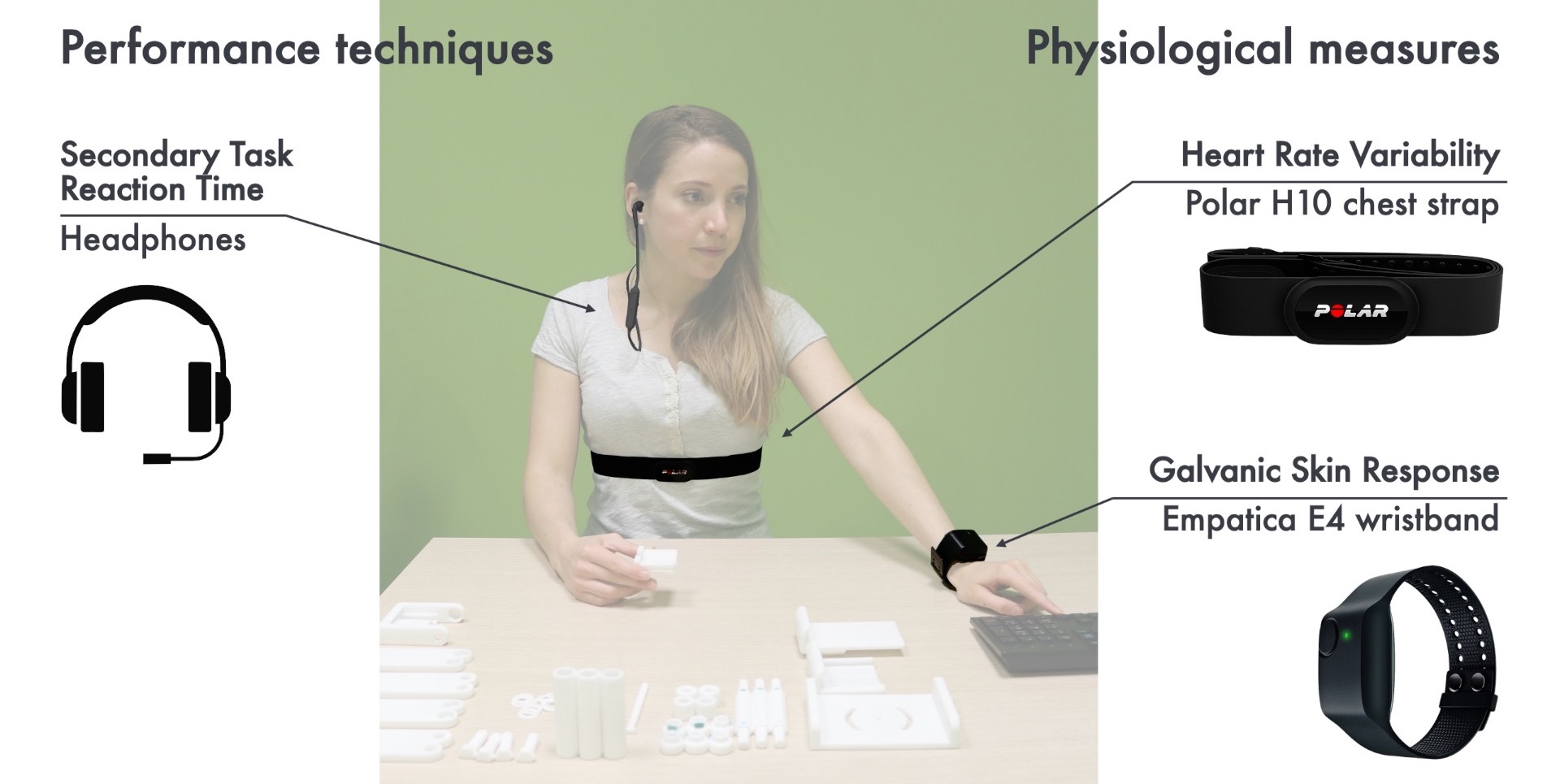}
    \caption{Measurements and sensors used as ground truth to test our metrics.}
    \label{fig:sensors}
\end{figure}

The experiments aimed to cross-test the performance of our cognitive load assessment framework against physiological measurements. 
In particular, the trend of the \textit{mental effort} was analysed in relation to heart rate variability, while the \textit{stress level} was compared with the commonly used features in galvanic skin response. The following section justifies the choice of these specific parameters as ground truth and describes the sensors adopted (also highlighted in Figure \ref{fig:sensors}) and the post-processing of the acquired signals.

\subsection{Baseline measurements}

\subsubsection{HRV responses: }
A chest strap (H10\footnote{\texttt{https://tinyurl.com/polarH10}}, Polar Electro Oy, Kempele, Finland) was used to record the electrocardiogram (ECG) signal. 
The RR interval, i.e. the time elapsed between two successive R-waves, 
were extracted from the raw ECG. 
Cardiovascular data analysis was subsequently performed using Kubios software\footnote{\texttt{https://www.kubios.com}}. 
The tool computes several classical metrics in time, frequency and non-linear domain. In this work, the frequency domain HRV data were considered. More precisely, the LF/HF ratio is selected since it is indicative of the mental effort, as suggested by the literature \citep{Mizuno2011, Durantin2014}. 

\subsubsection{Galvanic skin responses: }
The skin conductance was monitored by wristband Empatica E4\footnote{\texttt{https://www.empatica.com/en-eu/research/e4}}, a medical-grade wearable device acquiring real-time physiological data.
The recorded GSR signal was then processed using the open-source MATLAB toolbox Ledalab\footnote{\texttt{http://www.ledalab.de}}. 
A Butterworth low pass filter with a cut-off frequency at 2 Hz was used to filter the high-frequency components.
Finally, we applied the continuous decomposition analysis to separate the tonic (Skin Conductance Level, SCL) and phasic (Skin Conductance Response, SCR) components. 
As Marucci et al. \citep{Marucci2021}, we investigated the mean value of the SCL and the mean amplitude of the SCR peaks to assess the stress induced by the whole task on participants. 

\subsubsection{Secondary task performance: }
Concurrently with the primary assembly task, participants were asked some questions (three per experimental condition) through headphones. In the task-based methodology, performance on a secondary task is supposed to reflect the level of the cognitive load imposed by the primary task \citep{Paas2003}. 
We measured the reaction time of the user to the presented query whose answer is well known (e.g. the spelling of the name, the date of born, etc.).  
\subsubsection{Subjective questionnaire: }
At the end of the experiment, we asked participants to fill NASA-TLX \citep{Hart1988} and a custom questionnaire. The latter is a subjective scaling approach to capture mental effort- and stress-related factors in different task conditions. The evaluation includes a technique developed by NASA to assess the relative importance of factors in determining the experienced workload. Pairs of rating scale labels are presented, and the subject is asked to select which of the two was more relevant to the experience of cognitive workload in the task just performed. From the pattern of choices, we are able to associate a weight to each cognitive load factor and compute the overall score consistent with the experience of a specific subject.  
A copy of the custom questionnaire can be found as supplementary materials for this paper. 

\subsection{Experimental protocol}
The whole experimental procedure was carried out at Human-Robot Interfaces and Physical Interaction (HRII) Lab, Istituto Italiano di Tecnologia (IIT) in accordance with the Declaration of Helsinki, and the protocol was approved by the ethics committee Azienda Sanitaria Locale (ASL) Genovese N.3 (Protocol IIT\_HRII\_ERGOLEAN 156/2020).
All the subjects recruited were volunteers, naïve about the purpose of the experiment, and declared not to suffer from any mental disorder or cardiovascular disease.
The cognitive load employed by a worker is highly susceptible to the skills of the individual assessor. 
Thus, personnel without previous expertise and experience in the presented assembly task was considered. 

The study employed a within-subjects experiment in which each participant underwent all three experimental conditions. The tasks were devised with three levels of complexity (i.e. task 1 - simple, task 2 - medium, and task 3 - difficult) and industrial noise (i.e. task 1 - low, 45 dBA, task 2 - medium, 65 dBA, and task 3 - high, 75 dBA). The tasks order was defined as 1-2-3 for all the subjects, with the aim of imposing a growing complexity and thus identify an increase in cognitive effort.
The participants had fifteen minutes to complete each section. Before the beginning of the experiment, an initial calibration was performed to capture the physiological parameters and track the upper joints movements under resting conditions and then, set them as a reference. Moreover, the user had the chance to get familiar with the assembly parts and the interface for instructions.

The rest of the section describes two different experimental sessions that represent the consecutive phases in the development of our framework.

\subsection{Model calibration experiments}
The purpose of the first experimental session was to test the setup and identify the weights that should be associated with each cognitive load factor for the computation of cognitive load scores. To do this, five male subjects ($27.6 \pm 2.0$ years old) performed the whole experiment and filled in the custom questionnaire. 
Analysing factors' trend over time, we defined thresholds that thereafter permit the normalisation of the values assumed by cognitive load factors (i.e. $\in \,[0,1]$).
Given the patterns of choices in the questionnaire, we computed the weights that each subject would associate with each factor. The mean among all subjects for each factor weight 
was used in the second experimental session to create weighted combinations resulting in the \textit{mental effort} and \textit{stress level} scores.

\begin{figure*}[!htb]
    \centering
    \includegraphics[width=\linewidth]{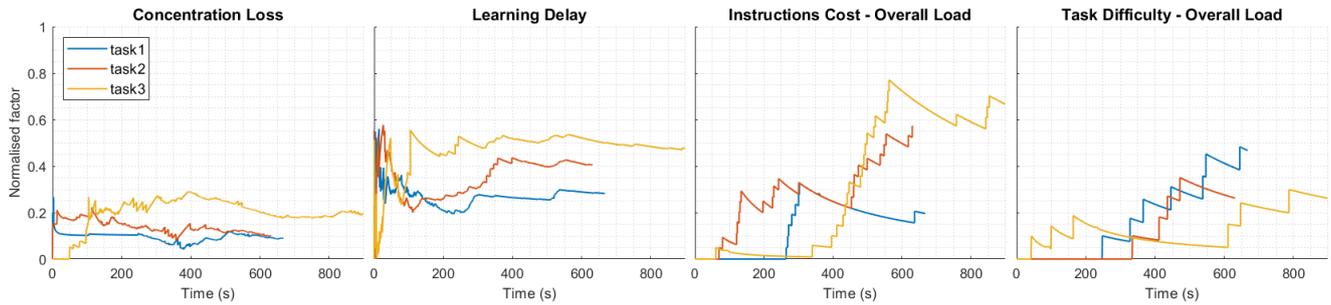}
    \caption{\textit{Concentration Loss}, \textit{Learning Delay}, \textit{Instructions Cost} and \textit{Task Difficulty} factors  associated to subject $1$ during three experimental conditions.}
    \label{fig:CL_factors}
\end{figure*}

\subsection{Multi-subject cognitive load assessment \\experiments}
Ten subjects, five males and five females ($26.6 \pm 3.7$ years old), were recruited for the second session. 
During the test, the cognitive load factors were computed online, and the final scores were shown on a monitor, only visible to the researcher (see Figure \ref{fig:experimentalSetup}). 
At the same time, physiological measurements were recorded. 
A statistical analysis was subsequently performed on the acquired data.
We adopted the non-parametric repeated measures Friedman's test to examine if the subject experienced different conditions imposed by the experiment (low, medium and high cognitive load). 
Finally, Spearman's rho correlation coefficient was used to assess if any relationship exists between the scores computed in the proposed framework and our ground truth measurements (i.e. physiological signals, performance measure and questionnaires).

\section{Experimental results} \label{results}

In this section, the results of the two experimental sessions are presented. We begin by outlining the outcomes of the model calibration experiments, highlighting the functioning of the final framework. 
This is followed by a deep analysis of cognitive load-related data acquired in multi-subject experiments.
Finally, we report the outcome of the online visual feedback interface.

\subsection{Model calibration}

\begin{table}[!b]
\caption{Thresholds and Weights associated with mental effort factors}
\resizebox{\linewidth}{!}{%
\begin{tabular}{@{}lrrr@{}}
\toprule
 & \multicolumn{2}{c}{Thresholds}                                  & \multicolumn{1}{c}{\multirow{2}{*}{Weights}} \\ \cmidrule(lr){2-3}
 & \multicolumn{1}{c}{Instantaneous} & \multicolumn{1}{c}{Overall} & \multicolumn{1}{c}{}                        \\ \midrule
\textit{Concentration Loss}     & -      & -     & 1.6 \\
\textit{Learning Delay}         & -      & -     & 3.2 \\
\textit{Concentration Demand}   & 12     & 26.0 & 1.6 \\
\textit{Instruction Cost}       & 13     & 26.1  & 4.0 \\
\textit{Task Difficulty}        & 6      & 10.7  & 2.2 \\
\textit{Frustration by Failure} & 2      & 4.7   & 3.0 \\
\textit{Tool Identification}    & -     & -     & 1.4 \\ \bottomrule
\end{tabular}
}
\label{tab:weights}
\end{table}

Table \ref{tab:weights} shows the results of the model calibration experiments. 
\textit{Concentration Loss} and \textit{Learning Delay} take on values between $0.0$ and $1.0$ by definition.  \textit{Tool Identification} factor saturates to $1.0$ after ten seconds as a practical choice. 
On the contrary, the other factors have to be normalised. 
To this aim, we defined upper-limit thresholds for each proposed factor as the maximum registered value for all subjects who took part in the first testing session. 

Besides, patterns of labels' choices in the custom questionnaire show the relative importance of the proposed factors in determining how much mental effort the operator is experiencing in the task. The third column of table \ref{tab:weights} illustrates the means of the weights given by participants to each factor. 
Interestingly, the cognitive demand to understand instructions (e.g. \textit{Instruction Cost}) represents the perceived most crucial contributor to workload.

\subsection{Cognitive load factor assessment}

\subsubsection{Mental effort: } 
Figure \ref{fig:CL_factors} displays the mental effort factors over time for one subject, as an example. 
We report \textit{Concentration Loss}, \textit{Learning Delay}, \textit{Instruction Cost} and \textit{Task Difficulty} factors since they show a meaningful trend throughout the task execution. On the other hand, the impact of \textit{Concentration Demand}, \textit{Frustration by Failure} and \textit{Tool Identification} factors is punctual when a specific event occurs.
The results of tasks 1, 2 and 3 are reported on the same chart to highlight differences in the trends. Note that the participant completed the first and second task before the total available time (i.e. fifteen minutes). 
The factors are normalised online, when needed, according to thresholds defined in the first experimental session.

The results obtained from the weighted combination of the factors are presented in Figure \ref{fig:mental_effort_vs_HRV} (first row). In particular, the black line sets out the trend of the \textit{mental effort} score over time in the different experimental conditions.    
Coloured bars highlight instead the score mean in three-minute intervals. 

\subsubsection{Stress level: } 
The estimated stress of a participant during the experiments is illustrated in Figure \ref{fig:stress_vs_GSR} (first row). Specifically, each occurrence of self-touching impacts 0.2 on the final score and as time passes, the contribution decreases (reaching zero after one minute).
The grey and black profiles represent hyperactivity and self-touching factors, respectively. By summing them, the \textit{stress level} score is evaluated and its mean within blocks lasting three minutes is reported as coloured bars. 

\begin{figure*}[!t]
    \centering
    \includegraphics[width=\linewidth]{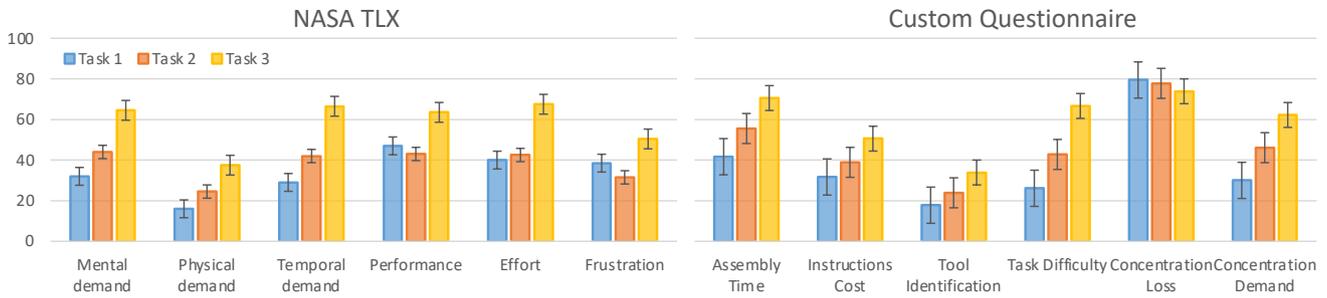}
    \caption{Results of the subjective evaluations (NASA TLX and custom questionnaire): bars represent mean and standard error between scores given by participants in different experimental conditions.}
    \label{fig:questionnaires}
\end{figure*}

\subsection{Significance of experimental conditions}

\subsubsection{Cognitive load scores: }
The \textit{mental effort} and \textit{stress level} means in three-minute intervals were compared with a Friedman's test to access differences in tasks across repeated measures.
The imposed conditions (i.e. increasing complexity and noise) affect the \textit{mental effort} score significantly, $\chi^2$=6.58, p=0.0373, and the \textit{stress level} score marginally, $\chi^2$=5.96, p=0.0507.
\subsubsection{HRV responses: }
Cardiovascular data analysis was performed on three-minute blocks of RR intervals of ECG signal. 
For all the subjects, we extract HRV features in frequency-domain and differences in their trend were assessed using Friedman's test with repeated measures (three-minute blocks). 
Different experimental conditions significantly impact the LF (Hz) parameter ($\chi^2$=6.68 p=0.0354), which exhibits a predominant decrease over the tasks. HF (Hz) showed instead marginal difference among the tasks ($\chi^2$=5.88 p=0.0529). 
Median LF/HF ratio levels for the low, medium and high imposed cognitive load experiments were 3.10, 3.64 and 3.77, respectively.
The statistical test revealed a significant difference in LF/HF ratio depending on the imposed complexity and noise, $\chi^2$=9.24 p=0.0098. 

\subsubsection{Galvanic skin responses: }
As Marucci et al. \citep{Marucci2021}, we investigated the mean value of the SCL and the mean amplitude of the SCR peaks during the experimental conditions. The analysis was performed in three-minute intervals with Friedman's test.
Both tonic and phasic components revealed a significant main effect of the load condition (p$<$0.01).

\subsubsection{Secondary task performance: } 
Friedman's test revealed a statistically significant difference in the reaction time of the secondary task ($\chi^2$=10.23 p$<$0.01). In general, the participants tended to delay the answer as the task complexity increases. 

\subsubsection{Subjective questionnaires: }

\begin{figure}[!b]
    {\begin{adjustwidth}{-0.1cm}{-0.3cm}
    \includegraphics[width=\linewidth]{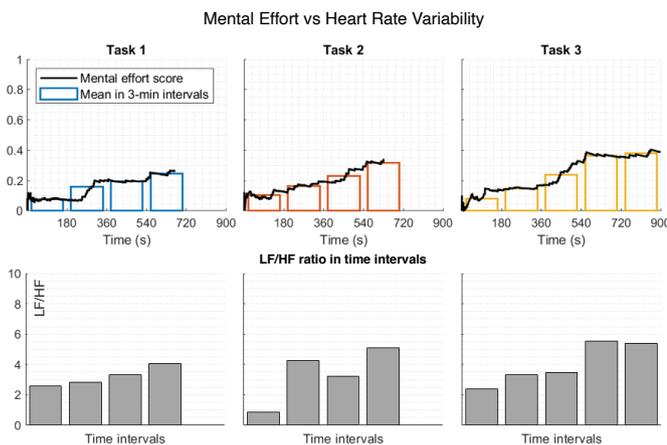}
    \end{adjustwidth}}
    \caption{Comparison between \textit{mental effort} score computed by our online framework and LF/HF ratio extracted from three-minute blocks of electrocardiography signal for subject $1$.}
    \label{fig:mental_effort_vs_HRV}
\end{figure}

\begin{figure}[!b]
    {\begin{adjustwidth}{-0.3cm}{-0.1cm}
    \vspace{0.4cm}
    \includegraphics[width=\linewidth]{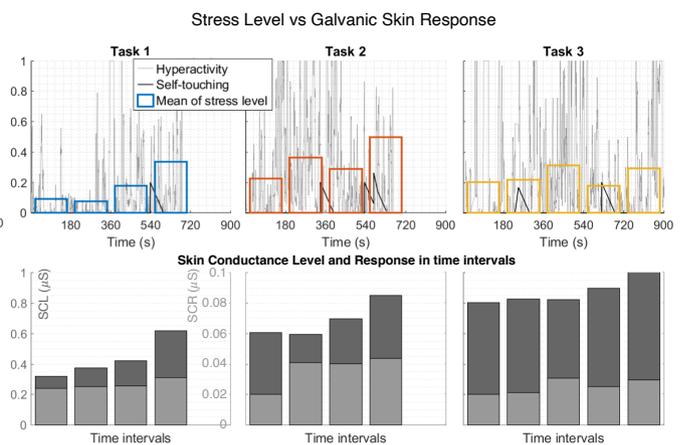}
    \end{adjustwidth}}
    \caption{Comparison between \textit{stress level} score and skin conductance level (SCL) and response (SCR) extracted from three-minute blocks of galvanic skin response for subject $1$.}
    \label{fig:stress_vs_GSR}
\end{figure}
Figure \ref{fig:questionnaires} presents the results of subjective questionnaires in the three experimental conditions. The bars represent the mean of the scores assigned by participants and the error bars display the 95\% confidence of scale means. The Kruskal-Wallis test was conducted to compare the three conditions in a systematic manner. The results from the NASA-TLX show statistically significant differences in mental demand ($\chi^2$=10.61 p=0.005) and effort ($\chi^2$=14.22 p=0.0008) among the tasks. For the other scores, the p-values were over 0.05 significance level.
From the custom questionnaire, we identify a significant effect in perceived concentration demand ($\chi^2$=7.11 p=0.0286), learning delay (i.e. automaticity in completing the assembly, $\chi^2$=6.48.11 p=0.0392) and task difficulty ($\chi^2$=14.33 p=0.0008) depending on imposed experimental conditions. 
Finally, the latter significantly affect the overall cognitive workload score ($\chi^2$=7.24 p=0.0267) computed by the ratings combination defined in the first testing session. 

\subsection{Correlation between cognitive load scores and physiological variables}

\begin{table*}[!t]
\caption{Spearman's correlation coefficients between estimated indicators of Cognitive Load (\textit{mental effort} and \textit{stress level}) and state-of-the-art measurements (physiological signals and performance) considering all participants.
}
\label{tab:correlations}
\centering
\begin{tabular}{@{}lrrrrrrrrrr@{}}
\toprule
& \multicolumn{10}{c}{Subject} 
\\ \cmidrule(l){2-11} 
 & \multicolumn{1}{c}{1} & \multicolumn{1}{c}{2} & \multicolumn{1}{c}{3} & \multicolumn{1}{c}{4} & \multicolumn{1}{c}{5} & \multicolumn{1}{c}{6} & \multicolumn{1}{c}{7} & \multicolumn{1}{c}{8} & \multicolumn{1}{c}{9} & \multicolumn{1}{c}{10} \\ \midrule
Mental effort/HRV & 0.80** & 0.74** & 0.62* & 0.67** & 0.60* & 0.87** & 0.49* & 0.20 & 0.06 & 0.21 \\
Mental effort/Secondary task & 0.42 & -0.46 & 0.53 & 0.82 & 0.47 & 0.47 & 0.03 & 0.60 & 0.26 & 0.54 \\ \midrule
Stress level/SCL & 0.43 & 0.58* & 0.63* & 0.72** & 0.07 & 0.01 & 0.47 & 0.31 & -0.13 & 0.01 \\
Stress level/SCR & 0.64* & 0.21 & 0.63* & 0.68* & -0.17 & 0.57* & 0.35 & 0.40 & 0.05 & -0.25 \\ \bottomrule
& \multicolumn{10}{l}{ Significance level are indicated at the *p\textless{}0.05, **p\textless{}0.005.}
\end{tabular}
\end{table*}

A Spearman's rank-order correlation was run to determine if a relationship exists between the scores computed in our framework and standard measures presented in the literature for the assessment of cognitive load. 
Figure \ref{fig:mental_effort_vs_HRV} compares the trend of the \textit{mental effort} score with the ratio of low-frequency power to high-frequency power (LF/HF ratio) in three-minute intervals extracted from ECG signals (second row). For seven out of ten participants, there was a strong, positive correlation between the mean within three-minute blocks of the computed score and the HRV feature, which was statistically significant (see Table \ref{tab:correlations} first row). 
For each subject, we also compute the correlation between the \textit{mental effort} score and the reaction time in the secondary task (three questions per experimental condition every three minutes). The test revealed a positive correlation, but no significance was found (see Table \ref{tab:correlations} second row).

The trend of the \textit{stress level} score is instead analysed in comparison with GSR-related measures. 
Figure \ref{fig:stress_vs_GSR} compares the trend of the \textit{stress level} score with the SCL and SCR features extracted by three-minute intervals of GSR signals (second row).
The bottom half of table \ref{tab:correlations} provides the summary statistics.  The skin conductance variables were partially correlated with the estimated score. 
In particular, positive correlations were detected, but they were statistically significant just for few subjects. 

\subsection{Online visual feedback}
\label{GUI}
\begin{figure}[!b]
    \centering
    \includegraphics[width=\linewidth]{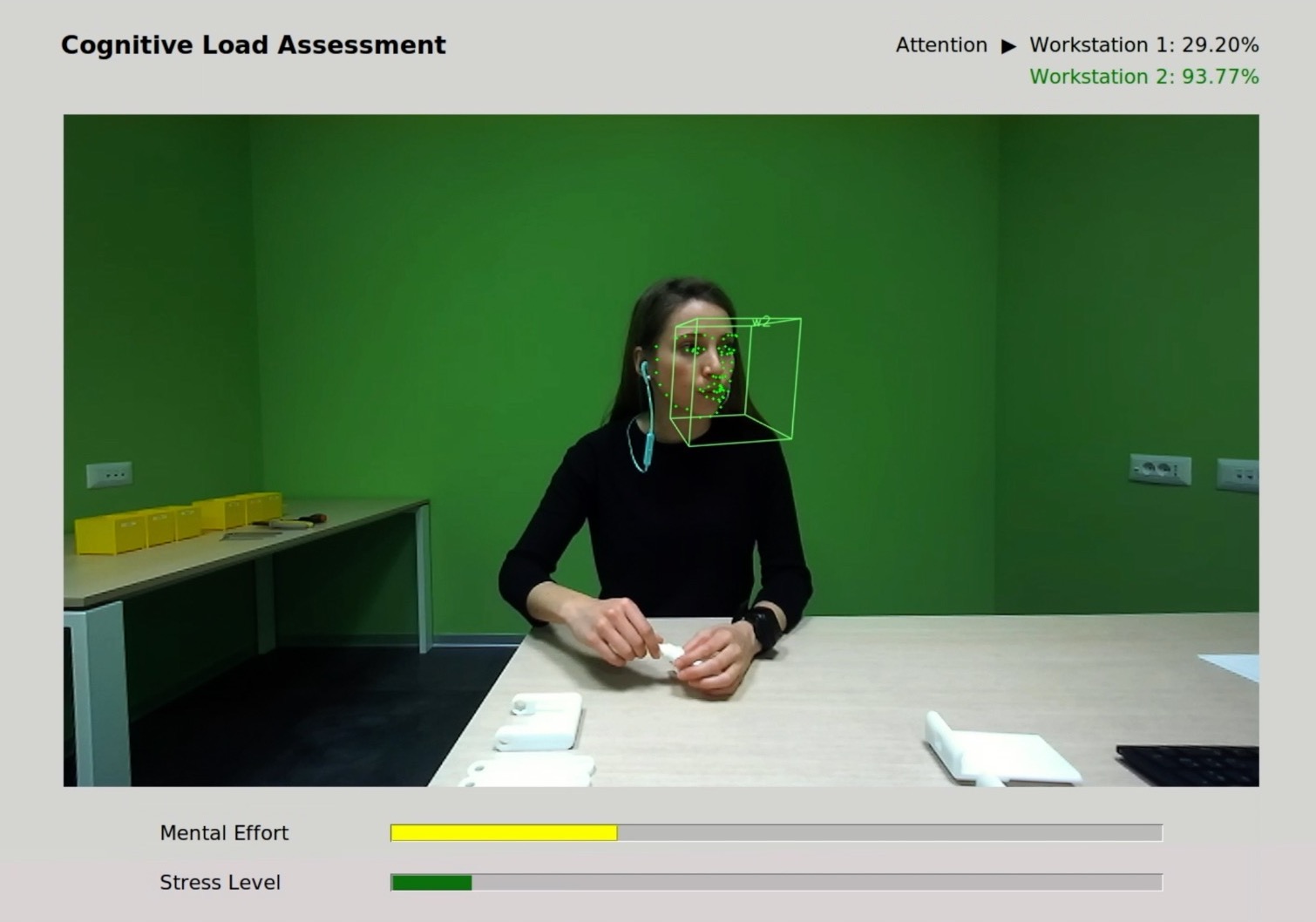}
    \caption{Online visual feedback reporting: current head direction, percentage of attention toward involved workstations, and estimated \textit{mental effort} and \textit{stress level} scores as colour-coded bars.}
    \label{fig:gui}
\end{figure}

Cognitive load scores are computed online since our final goal is to identify the excessive cognitive load and deliver warning messages to the human operator.
An interface was designed and implemented to provide visual feedback on the workload that the worker is currently experiencing during the assembly task (see Figure \ref{fig:gui}). 
The interface shows the real-time acquired video depicting the monitored subject. A pyramidal shape is drawn on the image to highlight the facing direction, and the percentage of attention toward the assembly ($W_1$) and instructions ($W_2$) workstations is specified.
Within this context, the workstation factors and \textit{instantaneous} parameters are considered, and their weighted combinations result in the instantaneous scores of \textit{mental effort} and \textit{stress level}. The latter is represented as colour-coded bars. Colour is used here to warn when excessive cognitive load is identified (green - low cognitive load, yellow - medium cognitive load, orange - high cognitive load, and red - very high cognitive load).
The reader can better understand the functioning of the implemented visual feedback by watching the video provided as supplementary material for this paper.

\section{Discussion}
\label{discussion}

From the performance comparison between our vision-based framework and state-of-the-art methods, 
we observed, in the trends, differences among the various experimental conditions (i.e. increasing task complexity and noise).
Therefore, statistical analysis was performed on the acquired data. 
Statistical significance through separate repeated-measures analysis of variance was found for both HRV and GSR features, 
as for the outcomes of the secondary task and questionnaires, in the different testing sessions.
Hence, we can infer that the subject actually experienced the imposed cognitive load conditions. 

A promising finding was that statistically significant differences were also identified in the cognitive load scores computed by our method (i.e. \textit{mental effort} and \textit{stress level}). 
This encouraged us to compare our online scores with state-of-the-art offline methods hardly deployable in industrial settings. 
As observed in Table \ref{tab:correlations}, the \textit{mental effort} mean in three-minute intervals appeared to be positively correlated to the LF/HF ratio extracted from ECG signal within the same time intervals and to secondary task performance. 
Moreover, positive correlations were detected between the \textit{stress level} and GSR-related features.

Results provided first evidence on the capability of the method to provide meaningful and reliable insights about the human cognitive load at work. 
Practical strong points of the setup include the reduced cost of the system components and the users' comfort while performing their tasks.
Our cognitive load assessment framework does not require the worker to wear any sensor and can be easily configured and used by non-experts in the areas of cognitive ergonomics and human-computer interaction. 

Besides, it is worth mentioning that the system is capable of identifying online excessive workload periods in assembly tasks and providing visual feedback using colour-coded bars.

\section{Conclusions}
\label{conclusions}

This paper presented an online and quantitative method to monitor the cognitive load of human operators by analysing the attention distribution and detecting motion patterns in assembly activities directly from the input images of a stereo camera. 
The main focus was on identifying risks in tasks and workstation design, where excessive workload might lead to errors or work difficulty.
We exploited cognitive load factors in manufacturing as identified by experts \citep{Thorvald2019} and evaluated them online through cutting-edge artificial intelligence algorithms (i.e. head pose estimation and skeleton tracking).
We estimated the \textit{mental effort} and \textit{stress level} currently experienced by the worker, investigating attention- and activity-related behavioural features, and we delivered intuitive warning messages on a screen. 

The proposed method shows promising features to be applied to the manufacturing sector. The framework indeed works online, does not require expensive equipment and does not ask the human worker to wear any sensor permitting the natural flow of work activities. 
The main limitation is the assumption of a well-structured working environment, where assembly instructions are shown on a monitor. A natural progression of this research is to generalise the framework, including more workstations and examining complicated industrial operations with multiple human workers.

Future works could also investigate the benefits of a subject-specific model of cognitive load processes to address individual demands and characteristics of the workers. 
Besides, both the weights and the thresholds defined for the developed factors could be tuned depending on each user feedback or previous user-specific calibration phase.  

The final results suggest that the presented method has the potential to be integrated into the development of human-robot interaction systems for improving human cognitive ergonomics in industrial settings.


\section*{Acknowledgment}
This work was supported in part by the ERC-StG Ergo-Lean (Grant Agreement No.850932), in part by the European Union’s Horizon 2020 research and innovation programme under Grant Agreement No. 871237 (SOPHIA).

\printcredits

\bibliographystyle{ieeetr}

\bibliography{bibliography}

\end{document}